\documentclass[11pt,a4paper]{article}
\usepackage[hyperref]{acl2019}
\usepackage{times}
\usepackage{latexsym}
\usepackage{url}
\usepackage{amsmath,graphicx}
\usepackage{url}
\usepackage{subfigure}
\usepackage{booktabs}
\usepackage{amssymb}
\usepackage{epsfig}
\usepackage{multirow}
\usepackage{hyperref}
\usepackage{color, colortbl}
\usepackage{footnote, tablefootnote}
\usepackage{verbatim}
\usepackage{blindtext}
\usepackage[utf8]{inputenc}

\aclfinalcopy 
\def\aclpaperid{1749} 

\title{SUMBT: Slot-Utterance Matching \\ for Universal and Scalable Belief Tracking}

\author{Hwaran Lee*
  \And Jinsik Lee* \\ SK T-Brain, AI Center, SK telecom \\ \texttt{ \{hwaran.lee, jinsik16.lee, oceanos\}@sktbrain.com }
  \And Tae-Yoon Kim}

\date{}

\begin{document}
\maketitle
\begin{abstract}
In goal-oriented dialog systems, belief trackers estimate the probability distribution of slot-values at every dialog turn.
Previous neural approaches have modeled domain- and slot-dependent belief trackers, and have difficulty in adding new slot-values, resulting in lack of flexibility of domain ontology configurations.
In this paper, we propose a new approach to universal and scalable belief tracker, called \textit{slot-utterance matching belief tracker} (SUMBT).
The model learns the relations between domain-slot-types and slot-values appearing in utterances through attention mechanisms based on contextual semantic vectors. Furthermore, the model predicts slot-value labels in a non-parametric way.
From our experiments on two dialog corpora, WOZ 2.0 and MultiWOZ, the proposed model showed performance improvement in comparison with slot-dependent methods and achieved the state-of-the-art joint accuracy.
\end{abstract}

{\let\thefootnote\relax\footnote{{*Hwaran Lee and Jinsik Lee equally contributed to this work.}}\addtocounter{footnote}{-1}}

\section{Introduction}\label{sec:intro}

As the prevalent use of conversational agents, goal-oriented systems have received increasing attention from both academia and industry. 
The goal-oriented systems help users to achieve goals such as making restaurant reservations or booking flights at the end of dialogs.
As the dialog progresses, the system is required to update a distribution over dialog states which consist of users' intent, informable slots, and requestable slots.
This is called belief tracking or dialog state tracking (DST).
For instance, for a given domain and slot-types (e.g., `restaurant' domain and `food' slot-type), it estimates the probability of corresponding slot-value candidates (e.g., `Korean' and `Modern European') that are pre-defined in a domain ontology.
Since the system uses the predicted outputs of DST to choose the next action based on a dialog policy, the accuracy of DST is crucial to improve the overall performance of the system. 
Moreover, dialog systems should be able to deal with newly added domains and slots\footnote{For example, as reported by \citet{kim-2018-alexa}, hundreds of new skills are added per week in personal assistant services.} in a flexible manner, and thus developing scalable dialog state trackers is inevitable.
Regarding to this, \citet{chen-2016-zeroshot} has proposed a model to capture relations from intent-utterance pairs for intent expansion. 

Traditional statistical belief trackers \cite{henderson-2014-delex} are vulnerable to lexical and morphological variations because they depend on manually constructed semantic dictionaries. With the rise of deep learning approaches, several neural belief trackers (NBT) have been proposed and improved the performance by learning semantic neural representations of words \cite{mrksic-2017-nbt, mrksic-2018-nbt}. However, the scalability still remains as a challenge; the previously proposed methods either individually model each domain and/or slot \cite{zhong-2018-glad, ren-2018-statenet, goel2018flexible} or have difficulty in adding new slot-values that are not defined in the ontology \cite{ramadan-2018-rnnnbt, nouri-2018-gce}.

In this paper, we focus on developing a ``scalable'' and ``universal'' belief tracker, whereby only a single belief tracker serves to handle any domain and slot-type.
To tackle this problem, we propose a new approach, called \textit{slot-utterance matching belief tracker} (SUMBT), which is a domain- and slot-independent belief tracker as shown in Figure \ref{fig:model}. 
Inspired by machine reading comprehension techniques \cite{chen2017reading, seo-2017-bidaf}, SUMBT considers a domain-slot-type (e.g., `restaurant-food') as a question and finds the corresponding slot-value in a pair of user and system utterances, assuming the desirable answer exists in the utterances. 
SUMBT encodes system and user utterances using recently proposed BERT \cite{devlin-2018-bert} which provides the contextualized semantic representation of sentences.
Moreover, the domain-slot-types and slot-values are also literally encoded by BERT. 
Then SUMBT learns \textit{the way where to attend} that is related to the domain-slot-type information among the utterance words based on their contextual semantic vectors. 
The model predicts the slot-value label in a non-parametric way based on a certain metric, which enables the model architecture not to structurally depend on domains and slot-types.
Consequently, a single SUMBT can deal with any pair of domain-slot-type and slot-value, and also can utilize shared knowledge among multiple domains and slots.

We will experimentally demonstrate the efficacy of the proposing model on two goal-oriented dialog corpora: WOZ 2.0 and MultiWOZ. We will also qualitatively analyze how the model works. Our implementation is open-published.\footnote{\url{https://github.com/SKTBrain/SUMBT}}

\section{SUMBT}\label{method}

\begin{figure}[!t]
    \centering
    \includegraphics[width=1\linewidth]{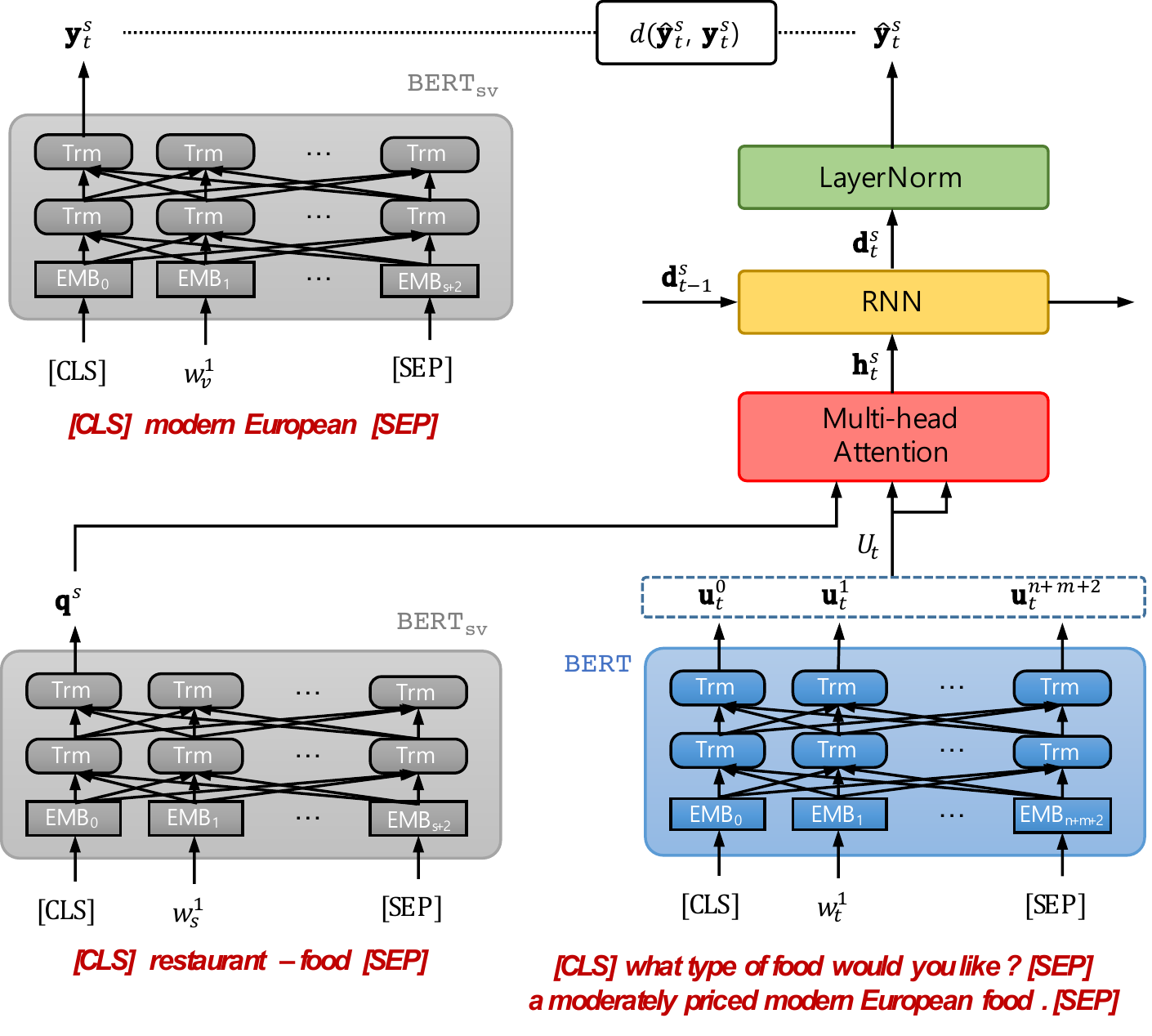}
    \caption{The architecture of slot-utterance matching belief tracker (SUMBT). An example of system and user utterances, a domain-slot-type, and a slot-value is denoted in red.
    }
    \label{fig:model}
\end{figure}

The proposed model consists of four parts as illustrated in Figure \ref{fig:model}: BERT encoders for encoding slots, values, and utterances (the grey and blue boxes); a slot-utterance matching network (the red box); a belief tracker (the orange box); and a non-parametric discriminator (the dashed line on top).

\subsection{Contextual Semantic Encoders}
For sentence encoders, we employed a pre-trained BERT model \cite{devlin-2018-bert} which is a deep stack of bi-directional Transformer encoders.
Rather than a static word vector, it provides effective contextual semantic word vectors. Moreover, it offers an aggregated representation of a word sequence such as a phrase and sentence, and therefore we can obtain an embedding vector of slot-types or slot-values that consist of multiple words.

The proposed method literally encodes words of domain-slot-types $s$ and slot-values $v_t$ at turn $t$ as well as the system and user utterances.
For the pair of system and user utterances, $\textbf{x}^{sys}_t = (w_1^{sys},...,w_n^{sys})$ and $\textbf{x}^{usr}_t = (w_1^{usr},...,w_m^{usr})$, the pre-trained BERT encodes each word $w$ into a contextual semantic word vector $\mathbf{u}$, and the encoded utterances are represented in the following matrix representation:
\begin{equation}
    U_t = \text{BERT}\left( \left[ \textbf{x}^{sys}_t, \textbf{x}^{usr}_t \right] \right).  
\end{equation}
Note that the sentence pairs are concatenated with a separation token $\text{[SEP]}$, and BERT will be fine-tuned with the loss function (Eq. \ref{eq:loss}).

For the domain-slot-type $s$ and slot-value $v_t$, another pre-trained BERT which is denoted as $\text{BERT}_\text{sv}$ encodes their word sequences $\mathbf{x}^s$ and $\mathbf{x}^v_t$ into contextual semantic vectors $\mathbf{q}^s$ and $\mathbf{y}^v_t$, respectively.
\begin{equation}
\begin{split}
    \mathbf{q}^s &= \text{BERT}_\text{sv}(\mathbf{x}^s), \\
    \mathbf{y}_t^v &= \text{BERT}_\text{sv}(\mathbf{x}^v_t).
\end{split}
\end{equation}
We use the output vectors corresponding to the classification embedding token $\text{[CLS]}$ that summarizes the whole input sequence.

Note that we consider $\mathbf{x}^s$ as a phrase of domain and slot words (e.g., $\mathbf{x}^s $ = ``restaurant -- price range'') so that $\mathbf{q}^s$ represents both domain and slot information.
Moreover, fixing the weights of $\text{BERT}_\text{sv}$ during training allows the model to maintain the encoded contextual vector of any new pairs of domain and slot-type.
Hence, simply by forwarding them into the slot-value encoder, the proposed model can be scalable to the new domains and slots.

\subsection{Slot-Utterance Matching}
In order to retrieve the relevant information corresponding to the domain-slot-type from the utterances, the model uses an attention mechanism. 
Considering the encoded vector of the domain-slot-type $\mathbf{q}^s$ as a query,
the model matches it to the contextual semantic vectors $\mathbf{u}$ at each word position, and then the attention scores are calculated. 

Here, we employed multi-head attention \cite{vaswani2017attention} for the attention mechanism.
The multi-head attention maps a query matrix $Q$, a key matrix $K$, and a value matrix $V$ with different linear $h$ projections, and then the scaled dot-product attention is performed on those matrices. 
The attended context vector $\mathbf{h}^s_t$ between the slot $s$ and the utterances at $t$ is 
\begin{equation}
    \mathbf{h}_t^s = \text{MultiHead}(Q, K, V),
\end{equation}
where $Q$ is $Q^s$ and $K$ and $V$ are $U_t$.

\subsection{Belief Tracker}

As the conversation progresses, the belief state at each turn is determined by the previous dialog history and the current dialog turn. The flow of dialog can be modeled by RNNs such as LSTM and GRU, or Transformer decoders (i.e., left-to-right uni-directional Transformer).

In this work, the attended context vector $\mathbf{h}_t$ is fed into an RNN,
\begin{equation}
    \mathbf{d}_t^s = \text{RNN}(\mathbf{d}_{t-1}^s, \mathbf{h}_t^s).
\end{equation}
It learns to output a vector that is close to the target slot-value's semantic vector.

Since the output of BERT is normalized by layer normalization \cite{ba2016layer}, the output of RNN $\mathbf{d}_t$ is also fed into a layer normalization layer to help training convergence,
\begin{equation}
    \hat{\mathbf{y}}_t^s = \text{LayerNorm}(\mathbf{d}_t^s).
\end{equation}

\subsection{Training Criteria}
The proposed model is trained to minimize the distance between outputs and target slot-value's semantic vectors under a certain distance metric. The probability distribution of a slot-value $v_t$ is calculated as 
\begin{equation}
\begin{split}
    p \left({v}_t |\mathbf{x}_{\leq t}^{sys}, \mathbf{x}_{\leq t}^{usr}, s \right)
    = 
    \frac{\exp \left(-d(\hat{\mathbf{y}}_t^s, \mathbf{y}_t^{v}) \right)}
    {\sum_{v' \in \mathcal{C}_s} \exp \left( -d(\hat{\mathbf{y}}_t^s, \mathbf{y}_t^{v'}) \right) },
    \end{split}
\end{equation}
where $d$ is a distance metric such as Euclidean distance or negative cosine distance, and $\mathcal{C}_s$ is a set of the candidate slot-values of slot-type $s$ which is defined in the ontology.
This discriminative classifier is similar to the metric learning method proposed in \citet{vinyals2016matching}, but the distance metric is measured in the fixed space that BERT represents rather than in a trainable space.

Finally, the model is trained to minimize the log likelihood for all dialog turns $t$ and slot-types $s\in \mathcal{D}$ as following:
\begin{equation} \label{eq:loss}
    \mathcal{L}(\mathbf{\theta}) = -\sum_{s \in \mathcal{D}}\sum_{t=1}^T{ \log p({v}_t |\mathbf{x}_{\leq t }^{sys}, \mathbf{x}_{\leq t} ^{usr}, s)}.
\end{equation}
By training all domain-slot-types together, the model can learn general relations between slot-types and slot-values, which helps to improve performance.

\section{Experimental Setup}
\subsection{Datasets}

To demonstrate the performance of our approach, we conducted experiments over WOZ 2.0 \cite{wen-2017-woz} and MultiWOZ \cite{budzianowski-2018-multiwoz} datasets. WOZ 2.0 dataset\footnote{Downloaded from \url{https://github.com/nmrksic/neural-belief-tracker}} is a single `restaurant reservation' domain, in which belief trackers estimate three slots (area, food, and price range).
MultiWOZ dataset\footnote{Downloaded from \url{http://dialogue.mi.eng.cam.ac.uk/index.php/corpus}. Before conducting experiments, we performed data cleansing such as correcting misspelled words.} is a multi-domain conversational corpus, in which the model has to estimate 35 slots of 7 domains. 

\subsection{Baselines}
\label{ssec:exp-baseline}

We designed three baseline models: \texttt{BERT+RNN}, \texttt{BERT+RNN+Ontology}, and a slot-dependent SUMBT. 1) The \texttt{BERT+RNN} consists of a contextual semantic encoder (\texttt{BERT}), an RNN-based belief tracker (\texttt{RNN}), and a linear layer followed by a softmax output layer for slot-value classification. The contextual semantic encoder in this model outputs aggregated output vectors like those of $\text{BERT}_\text{sv}$. 2) The \texttt{BERT+RNN+Ontology} consists of all components in the \texttt{BERT+RNN}, an ontology encoder (\texttt{Ontology}), and an ontology-utterance matching network which performs element-wise multiplications between the encoded ontology and utterances as in \citet{ramadan-2018-rnnnbt}. Note that two aforementioned models \texttt{BERT+RNN} and \texttt{BERT+RNN+Ontology} use the linear layer to transform a hidden vector to an output vector, which depends on a candidate slot-value list. In other words, the models require re-training if the ontology is changed, which implies that these models have lack of scalability. 3) The slot-dependent SUMBT has the same architecture with the proposed model, but the only difference is that the model is individually trained for each slot.

\subsection{Configurations}
We employed the pre-trained BERT model that has 12 layers of 784 hidden units and 12 self-attention heads.\footnote{The pretrained model is published in \\ \url{https://github.com/huggingface/pytorch-pretrained-BERT}}
We experimentally found the best configuration of hyper-parameters in which search space is denoted in the following braces.
For slot and utterance matching, we used the multi-head attention with $ \left\{ 4, 8 \right\}$ heads and 784 hidden units.
We employed a single-layer $\{\text{GRU},\text{LSTM}\}$ with $\{100, 200, 300\}$ hidden units as the RNN belief tracker. 
For distance measure, both Euclidean and negative cosine distances were investigated.
The model was trained with Adam optimizer in which learning rate linearly increased in the warm-up phase then linearly decreased. 
We set the warm-up proportion to be $\{0.01, 0.05, 0.1\}$ of $\{300, 500\}$ epochs and the learning rate to be $\{1 \times 10^{-5}, 5 \times 10^{-5} \}$. The training stopped early when the validation loss was not improved for 20 consecutive epochs.
We report the mean and standard deviation of joint goal accuracies over 20 different random seeds.
For reproducibility, we publish our PyTorch implementation code and the pre-processed MultiWOZ dataset.
\section{Experimental Results}

\subsection{Joint Accuracy Performance}
The experimental results on WOZ 2.0 corpus are presented in Table \ref{tab:woz}.
The joint accuracy of SUMBT is compared with those of the baseline models that are described in Section \ref{ssec:exp-baseline} as well as previously proposed models. 
The models incorporating the contextual semantic encoder BERT beat all previous models. 
Furthermore, the three baseline models, \texttt{BERT+RNN}, \texttt{BERT+RNN+Ontology}, and the slot-dependent SUMBT, showed no significant performance differences. On the other hand, the slot-independent SUMBT which learned the shared information from all across domains and slots significantly outperformed those baselines, resulting in 91.0\% joint accuracy. This implies the importance of utilizing common knowledge through a single model. 

Table \ref{tab:multiwoz} shows the experimental results of the slot-independent SUMBT model on MultiWOZ corpus. Note that MultiWOZ has more domains and slots to be learned than WOZ 2.0 corpus. The SUMBT greatly surpassed the performances of previous approaches by yielding 42.4\% joint accuracy. The proposed model achieved state-of-the-art performance in both WOZ 2.0 and MultiWOZ datasets.

\begin{table}[t]
  \small
 {\renewcommand{\arraystretch}{1.3} 
    \begin{tabular}{ll}
    \hline
    Model & \multicolumn{0}{l}{Joint Accuracy} \\
    \hline
    NBT-DNN \cite{mrksic-2017-nbt} & $0.844$ \\
    BT-CNN \cite{ramadan-2018-rnnnbt} & $0.855$ \\
    GLAD \cite{zhong-2018-glad} & $0.881$ \\
    GCE \cite{nouri-2018-gce} & $0.885$ \\
    StateNetPSI \cite{ren-2018-statenet} & $0.889$ \\
    \hline
    BERT+RNN (baseline 1) & $0.892\ (\pm 0.011)$ \\
    BERT+RNN+Ontology (baseline 2) & $0.893\ (\pm 0.013)$ \\
    Slot-dependent SUMBT (baseline 3) & $0.891\ (\pm 0.010)$ \\
    Slot-independent SUMBT (proposed) & $\textbf{0.910}\ (\pm 0.010)$ \\
    \hline
    \end{tabular}%
    }
    \caption{Joint goal accuracy on the evaluation dataset of WOZ 2.0 corpus.}
  \label{tab:woz}%
\end{table}%

\begin{table}[t]
  \centering
  \small
 {\renewcommand{\arraystretch}{1.3} 
    \begin{tabular}{lc}
    \hline
    Model & Joint Accuracy \\
    \hline
    Benchmark baseline \footnotemark
    & \multicolumn{1}{l}{$0.2583$} \\
    GLAD \cite{zhong-2018-glad} & \multicolumn{1}{l}{$0.3557$} \\
    GCE \cite{nouri-2018-gce} & \multicolumn{1}{l}{$0.3558$} \\
    \hline
    SUMBT  & $\textbf{0.4240}\ (\pm 0.0187)$ \\
    \hline
    \end{tabular}%
    }
  \caption{Joint goal accuracy on the evaluation dataset of MultiWOZ corpus.}
  \label{tab:multiwoz}%
\end{table}%
\footnotetext{\small{
The benchmark baseline is the model proposed in \citet{ramadan-2018-rnnnbt} and the performance is described in \url{http://dialogue.mi.eng.cam.ac.uk/index.php/corpus/}
    }
}

\subsection{Attention Weights Analysis} \label{ssec:attn}
\begin{figure*}[t]
    \centering
    \includegraphics[width=0.8\linewidth]{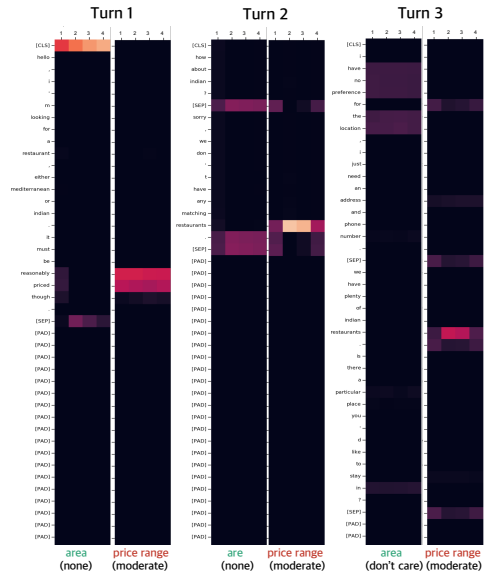}
    \caption{Attention visualizations of the first three turns in a dialog (WOZ 2.0).
    At each turn, the first and second columns are the attention weights when the given slots are `area' and `price range', respectively. The slot-value labels are denoted in the parentheses.}
    \label{fig:attention}
\end{figure*}

Figure \ref{fig:attention} shows an example of attention weights as a dialog progresses.
We can find that the model attends to the part of utterances which are semantically related to the given slots, even though the slot-value labels are not expressed in the lexically same way.
For example, in case of `price range' slot-type at the first turn, the slot-value label is `moderate' but the attention weights are relatively high on the phrase `reasonably priced'.
When appropriate slot-values corresponding to the given slot-type are absent (i.e., the label is `none'), the model attends to $\text{[CLS]}$ or $\text{[SEP]}$ tokens.

\section{Conclusion}
In this paper, we propose a new approach to universal and scalable belief tracker, called SUMBT which attends to words in utterances that are relevant to a given domain-slot-type. Besides, the contextual semantic encoders and the non-parametric discriminator enable a single SUMBT to deal with multiple domains and slot-types without increasing model size. The proposed model achieved the state-of-the-art joint accuracy performance in WOZ 2.0 and MultiWOZ corpora. Furthermore, we experimentally showed that sharing knowledge by learning from multiple domain data helps to improve performance.
As future work, we plan to explore whether SUMBT can continually learn new knowledge when domain ontology is updated.

\section*{Acknowledgements}
We would like to thank Jinyoung Yeo and anonymous reviewers for their constructive feedback and helpful discussions.
We are also grateful to SK T-Brain Meta AI team for GPU cluster supports to conduct massive experiments.

\bibliography{acl2019}
\bibliographystyle{acl_natbib}

\end{document}